\documentclass[10pt,twocolumn,letterpaper]{article}

\usepackage{iccv}              
\usepackage{multirow}
\usepackage{enumitem}
\setlist[description]{font=\normalfont\bfseries,leftmargin=2cm,labelsep=1em}
\usepackage{float}
\usepackage{cuted}
\usepackage[section]{placeins}

\definecolor{iccvblue}{rgb}{0.21,0.49,0.74}
\usepackage[pagebackref,breaklinks,colorlinks,allcolors=iccvblue]{hyperref}

\def\paperID{*****} 
\def\confName{ICCV}
\def\confYear{2025}

\title{Understanding Dataset Bias in Medical Imaging: A Case Study on Chest X-rays}
\author{Ethan Dack\\
University of Bern \\
{\tt\small ethan.dack@unibe.ch}
\and
Chengliang Dai\\
UCB Biopharma UK \\
Imperial College London \\
{\tt\small c.dai@imperial.ac.uk}
}

\begin{document}
\maketitle
\begin{abstract}
Recent works \cite{liu2024decade, zengyin2024bias} have revisited the infamous task ``Name That Dataset'' \cite{5995347}, demonstrating that non-medical datasets contain underlying biases and that the dataset origin task can be solved with high accuracy.
In this work, we revisit the same task applied to popular open-source chest X-ray datasets.
Medical images are naturally more difficult to release for open-source due to their sensitive nature, which has led to certain open-source datasets being extremely popular for research purposes.
By performing the same task, we wish to explore whether dataset bias also exists in these datasets.
To extend our work, we apply simple transformations to the datasets, repeat the same task, and perform an analysis to identify and explain any detected biases.
Given the importance of AI applications in medical imaging, it's vital to establish whether modern methods are taking shortcuts or are focused on the relevant pathology. 
We implement a range of different network architectures on the datasets: NIH \cite{wang2017chestx}, CheXpert \cite{irvin2019chexpert}, MIMIC-CXR \cite{johnson2019mimic} and PadChest \cite{bustos2020padchest}. 
We hope this work will encourage more explainable research being performed in medical imaging and the creation of more open-source datasets in the medical domain.
Our code can be found here: \href{https://github.com/eedack01/x_ray_ds_bias}{https://github.com/eedack01/x\_ray\_ds\_bias}.
\end{abstract}
\section{Introduction}
\label{sec:intro}


\begin{figure}[t]
\centering
\includegraphics[width=\linewidth]{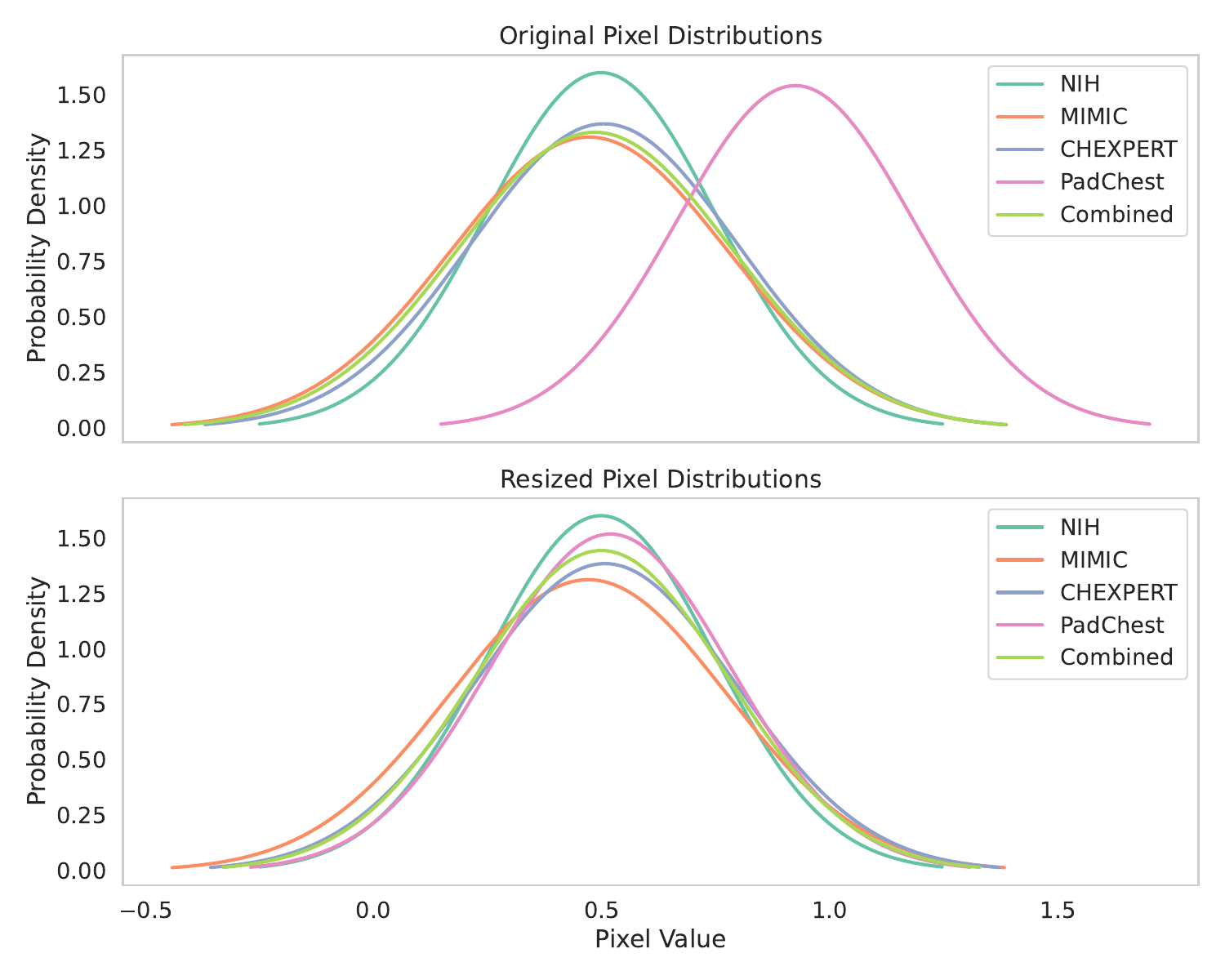}
\caption{We computed the distributions of selected datasets before and after data preparation. The PadChest distribution prior to resizing differs substantially due to its smaller image size compared to the other datasets. After resizing all datasets to a common resolution, the pixel value distributions are quite similar.}
\label{fig:pixel_dists}
\end{figure}

Working with in-house medical data is often challenging due to the extensive pre-processing, cleaning, and standardisation required before analysis \cite{deheyab2022future, PESAPANE2025111852}. 
Moreover, bias can be introduced at any stage of the pipeline—from data collection and labelling to preprocessing, model training, and evaluation—making it critical to carefully audit each step to ensure data integrity and fairness.
Furthermore, if the in-house data is not released at a later date, any research will not be replicable by other unaffiliated institutions \cite{Colliot2023}.
Any models developed on the in-house data alone should be further validated on external data to ensure no bias exists \cite{PESAPANE2025111852}.
One example of clear bias can be seen in machine learning methods that focus on the abundance of COVID-19 imaging data collected during the pandemic.
Although there were many research publications with papers often achieving over 99\% accuracy, these models were more than often deemed unusable in clinical practice \cite{roberts2021common}.
A possible explanation is that the patient splits may not have been conducted correctly; if a patient had multiple images appearing in both the training and validation sets, the model could easily minimise losses by effectively memorising patient-specific features.
To negate these issues, the alternative approach is to train models with large, diverse, open-source datasets \cite{ritore2024role}.
Not only is the research reproducible, but the data is often much easier to work with than in-house data.
Based on this, we focus our analysis on dataset bias in widely used open-source X-ray datasets for deep learning research.
An obvious cause of bias is the origin of the data, for instance, more than half of the clinical datasets used for developing and validating AI algorithms originated from either the United States (US) or China \cite{celi2022sources}.
Another instance would be the machine manufacturer, as studied in \cite{biondetti2020name}, who predicts the manufacturing origin.
However, these biases should be mitigated by sensible pre-processing steps \cite{rouzrokh2022mitigating}. 
The datasets chosen in this study have received an enormous amount of attention in deep learning research since their release, amounting to over 10,000 citations. 
As these datasets are much more straightforward to use when compared to in-house data, they are often the subject of self-supervised learning approaches, resulting in foundation models.
Such dataset bias is hopefully avoided in the large open-source datasets, but because of the high number of dimensions which occur in deep learning, it is difficult to avoid all bias \cite{glocker2023risk}. 
In some of the datasets, imbalanced classes do occur, which is natural due to the occurrence of rare diseases, but deep learning models will take shortcuts often due to the metrics they are trying to solve \cite{mosquera2024class}. 
Initial work in 2011 that first proposed the task \textit{Name That Dataset!} was able to achieve rather surprising results with traditional machine learning methods. 
Within the last year, this task has been popularised by recent works repeating the same task with much larger and more diverse datasets, along with more sophisticated modern models. 
We analyse the chosen open-source X-ray datasets by replicating these experiments.
This task is quite different due to the nature of medical images.
X-ray images only contain a single channel, whereas natural images contain three.
Natural image distributions over the RGB channels can often vary much more, as seen in \cite{zengyin2024bias}. 
The distributions are very similar in Figure \ref{fig:pixel_dists}, showing that models should struggle to predict the dataset origin, \textit{Will this make the task more difficult?}
We emulate a mixture of recent studies, which goes beyond the \textit{Name that dataset} paper by performing experiments on semantic structures and contour images.
By replicating experiments using a range of different models, this work successfully shows that dataset bias can exist even in carefully curated, large, open-source medical datasets.

\section{Related Work}
\label{sec:related_work}

\paragraph{Open-source X-ray datasets} have greatly contributed to progress in AI-driven medical research.
There have been many different X-ray datasets released in the last few years \cite{demner-fushman2016radiology, majkowska2019chest, nguyen2020vindr}.
We observe that there have been datasets originating that focus on certain diseases \cite{jaeger2014two, shih2019augmenting, filice2020crowdsourcing, cohen2021radiographic}.
Albeit with the large amount of choice, researchers often focus on the easiest to access and the largest in quantity, due to the importance of having a large amount of data, the specialised medical institute's datasets are often smaller in size.
We opted to select the four largest available datasets.
As some of the datasets contain different angles, we decided to choose the Frontal position, which exists in all datasets; this resulted in a total of 655,176 Frontal images.
\textbf{"ChestX-ray8"} \cite{wang2017chestx}, commonly denoted as \textbf{NIH}, consists of only frontal scans from 32,717 unique patients with eight common thoracic diseases. 
\textbf{CheXpert} \cite{irvin2019chexpert}, contains multiple image angles and consists of 65,240 patients; within this study, a labeller was carefully designed to choose one of fourteen labels.
\textbf{MIMIC-CXR} \cite{johnson2019mimic}, we will shorten to \textbf{MIMIC}, contains images from 65,379 patients, labelling was also done by the same labeller as in \textbf{CheXpert}.
\textbf{PadChest} \cite{bustos2020padchest} includes images from more than 67,000 patients, 19 different diagnoses and contains the largest amount of manually annotated data of the datasets.
All of these datasets made use of automated labelling via text mining methods, proposing three different methods.
Whilst this is efficient, there is a small amount of error in the labels of the dataset.
\textbf{NIH}, \textbf{CheXpert} and \textbf{MIMIC} were all sourced from the USA whereas \textbf{PadChest} was sourced from Spain.
Since the release of these datasets, there has not been such a major dataset released which matches in size.
The labels in each dataset are overlapping, with \textbf{NIH} being a subset of \textbf{CheXpert} and \textbf{MIMIC}, and all three datasets being a subset of \textbf{PadChest}.
These overlapping labels are crucial; if all images of a dataset belonged to one disease class and all images of another belonged to a different disease class, this would be trivial supervised learning for models.

\paragraph{Dataset Bias} is not to be confused with `data' bias.
In this work, the datasets involved are sampling real-world data for specific medical applications. 
This is not surprising that dataset bias can be considered a major area of interest when looking at images that cover the same area \cite{tommasi2017deeper}. 
Torralba and Efros \cite{5995347} popularised questioning open-source datasets by doing two simple tasks: \textit{Name That Dataset!} and Cross dataset generalisation. 
The first task involves sampling images from different datasets and training a classifier to predict the images' origin.
Cross-dataset generalisation is training a model for a task on one dataset and evaluating its performance on a different dataset.
Cross-dataset generalisation is not our focus due to medical diagnosis algorithms regularly being validated on different datasets, for example, in the foundation model \cite{Tiu2022}.
Capture bias, a common source of dataset bias, arises when the object of interest is consistently positioned at the centre of the image \cite{Fabbrizzi2022, 37648}.
This is especially relevant in X-rays, where the heart and lungs occupy the central region.
Khosla et al \cite{khosla2012undoing} propose an interesting solution for dealing with bias when merging multiple datasets by predicting two sets of weights: bias vectors associated with each individual dataset and a set of visual world weights that unifies features for all datasets. 
Tommasi et al \cite{tommasi2017deeper} propose to validate features and a new metric to evaluate the ability of a model to address dataset bias. 
This analysis aims to provide a deeper understanding of what dataset bias occurs in large open-source X-ray datasets.

\begin{figure}[t]
\centering
\includegraphics[width=\linewidth]{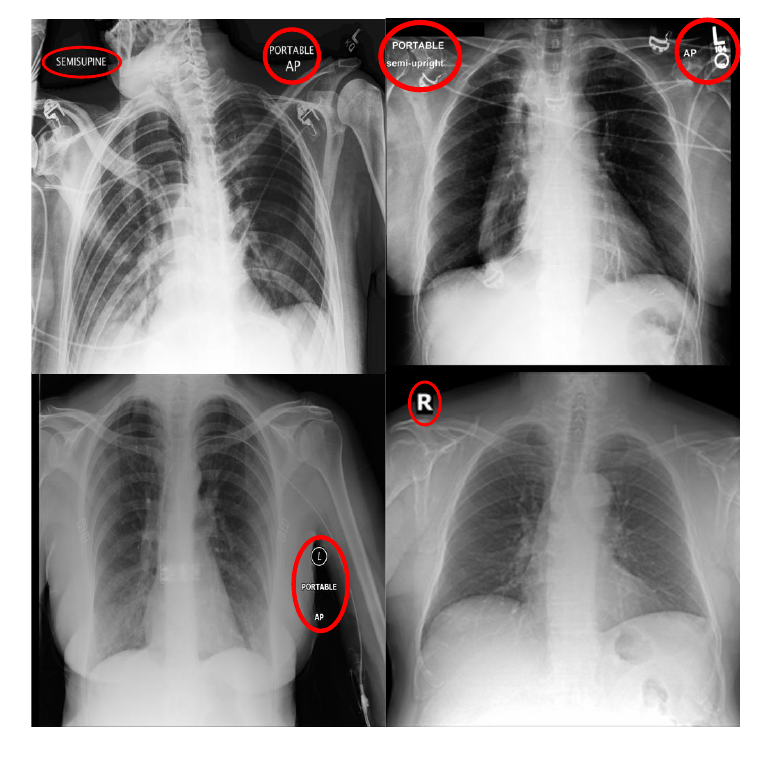}
\caption{Choosing a random image from each dataset, we can clearly observe possible bias artefacts present in each dataset}
\label{fig:artefacts}
\end{figure}

\paragraph{Bias in Deep learning with X-rays} has been examined in several studies highlighting bias issues in COVID-19 and Pneumonia X-ray datasets \cite{GARCIASANTACRUZ2021102225, ariasgarzon2023biases, catala2021bias}. 
We take influence from \cite{catala2021bias} by making use of heatmaps to understand how our trained models classify the dataset origin.
An obvious reason for bias in X-ray imaging could be due to man-made artefacts as shown in Figure \ref{fig:artefacts}.
Artefacts do not only include these, but can refer to any unclear appearance of anatomical structure due to random error in radiographic techniques.
These artefacts or confounding features can be particularly problematic when training models for diagnosis, for instance, we see deep learning models will take shortcuts when trained to differentiate between COVID-19 and pneumonia \cite{benahmed2021deep}. 
Such feature shortcuts can be classified as spurious correlations \cite{Simon1977, jones2024causal}.
Several studies have analysed the presence of spurious correlations and attempted to negate them \cite{Hagos2023, Kumar2023}.
Additional correlations can be caused by incorporating external demographic features in disease classification.
Research has found that models with less encoding of demographic features exhibited more fairness \cite{Yang2024}.
Still, even larger studies analysing larger models have shown that demographic data is key to identifying bias \cite{Vaidya2024,yang2024demographic}. 
Whilst this is true, such demographics need to be carefully curated whilst creating open-source datasets, and thus are not always available.
In this study, we opt to focus on the imaging data alone to analyse whether models can identify bias.

\section{Identifying the Bias}

If we take the images as they are and do minimal preparation, it would be extremely trivial for modern neural networks to solve the task we analyse in this paper.
To this end, we perform data preparation and various transforms in our experiments to develop a deeper understanding of how our models can classify their origins.
We are trying to minimise the effect of artefacts, and we wish models to identify pathologies or even identify correlations present in each dataset.

\subsection{Data preparation}

We obtain all the datasets chosen from their respective providers except for \textbf{PadChest}, which had been converted into images from DICOM by TorchXRayVision \cite{Cohen2022xrv}.
Avoiding low-level signatures is often uncanny to the human eye but could be easily identified by neural networks \cite{liu2024decade}.
To this extent, we undergo some basic preprocessing to keep datasets in similar distributions.
First, we ensure all datasets are stored in a JPG format with the same compression factor; \textbf{NIH} and \textbf{PadChest} were stored in PNG, so we convert these.
Each dataset consists of a different number of images, so class imbalance would occur if we tried to use the maximum number of images. 
To counter this, we decide to sample 100,000 from each dataset to ensure balance.
The datasets also consist of different image sizes, so we resize all images to be stored as (512, 512) to keep pixel value distributions similar.
We aim to make the data format as similar as possible, which keeps fairness before any experiments begin. 
As stated, we do not make use of any available metadata as different metadata exists for each dataset; we only make use of Patient identification numbers to split data between train and test, i.e. patient images cannot exist in both the train and test sets.


\begin{figure*}[t]
\centering
\includegraphics[width=\textwidth]{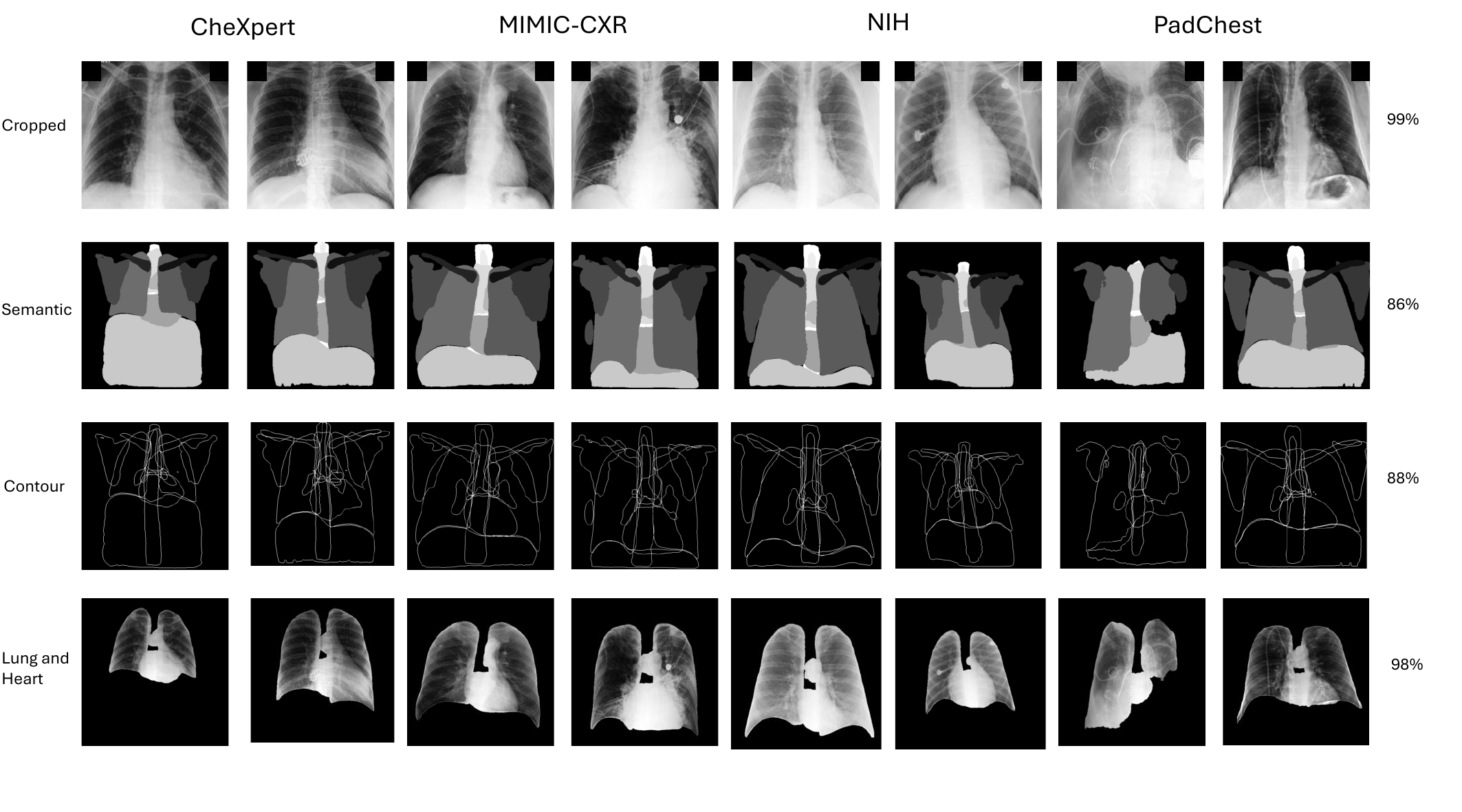}
\caption{As in \cite{zengyin2024bias}, we apply various transformations to the sampled images, selecting two random images from each dataset for each transformation. The first row shows the cropped images with black patches; the second row displays the corresponding semantic maps; the third row shows the contours; and the final row presents the combined lung and heart images alone.}
\label{fig:transforms}
\end{figure*}

\subsection{Chosen Models}
Since the widespread adoption of deep learning, many architectures have emerged. 
We select three well-established models spanning the period from 2011 to 2021 to demonstrate that dataset bias exists regardless of model choice. 
For all model inputs, we load the image as grayscale so the number of channels is equal to one.
All models had a final fully connected (linear) layer in the classifier that was replaced to output predictions for the 4-way classification task.

\paragraph{AlexNet}\cite{NIPS2012_c399862d}, consisting of 61,100,840 parameters, proposed in 2012, popularised image classification with convolutional neural networks (CNNs).
It achieved the highest result on the ImageNet task \cite{5206848} by a large distance. 
To allow the training of images of a single channel, we modify the first convolutional layer to accept 1 input channel (grayscale, e.g., for X-ray images) instead of the original 3 channels (RGB).
As the field is moving so quickly now, many alternative CNNs have been released. 
AlexNet was prone to classic problems, such as a vanishing gradient.
Nonetheless, we believe that as the model had such an impact, we believe it is a valid model to evaluate in terms of complexity.

\paragraph{ResNet50} \cite{resnet50}, released in 2015, containing 25,557,032 parameters, significantly fewer parameters than AlexNet, but its efficient architecture will often outperform AlexNet with ease. 
The introduction of residual connections solved the problems with the vanishing gradient when optimisers update the weights of the network.
As implementation, we make use of the open-source MONAI library \cite{cardoso2022monaiopensourceframeworkdeep}, which allows us to modify the spatial inputs, channels and classes with ease.

\paragraph{Vision Transformer} (ViT) \cite{dosovitskiy2020vit}, the newest model we evaluate, released in 2021, is made up of 86,567,656 parameters, which is expected due to the extra calculation steps performed in ViTs.
This work popularised transformers for images by splitting the image up into separate patches, whilst achieving similar results to ResNet, it is often more unstable in training \cite{steiner2022trainvitdataaugmentation}.
We evaluate the ViT due to its recent popularity for its use as an encoder for self-supervised training \cite{Tiu2022, xiao2023delving}.
We also implemented using the MONAI library for the previous ResNet50 reasons.

\begin{table*}[t]
\caption{F1 scores per dataset for each classification task. ResNet50 consistently achieved the highest performance across all tasks, while ViT had the lowest. Among the datasets, PadChest yielded the highest F1 scores, whereas MIMIC or NIH typically had the lowest.}
\label{tab:combined_results}
\centering
\setlength{\tabcolsep}{12pt}
\begin{tabular}{ll|ccccc}
\toprule
\textbf{Image Type} & \textbf{Model} & \textbf{CheXpert} & \textbf{MIMIC} & \textbf{NIH} & \textbf{PadChest} & \textbf{Average}\\
\midrule

\multirow{3}{*}{Cropped Images} 
& AlexNet    & 98.35 & 97.34 & 97.27 & 98.30 & 97.82  \\
& ResNet50  & 99.28  & 98.79 & 99.35 & 99.70 & \textbf{99.28} \\
& Vit-B-16   & 95.86  & 95.13 & 94.79  &95.94 & 95.43 \\
& \textbf{Average} & 97.83  & 97.01 & 97.14  & \textbf{97.98} & 97.51 \\

\midrule

\multirow{3}{*}{Semantic Images} 
& AlexNet     & 84.66 & 81.97 & 83.52 & 91.80 & 85.49\\
& ResNet50  & 86.38  & 82.57 & 84.90 & 92.53 & \textbf{86.59} \\
& Vit-B-16    & 70.21  & 65.94 & 60.54 & 76.78 & 68.37 \\
& \textbf{Average}   & 80.42  & 76.83 & 76.32  & \textbf{87.04} & 80.15 \\

\midrule

\multirow{3}{*}{Contour Images} 
& AlexNet   & 84.77 & 82.77 & 84.90 & 92.66 & 86.27 \\
& ResNet50  & 87.96 & 85.15 & 87.47 & 93.91 & \textbf{88.62}  \\
& Vit-B-16  & 64.92 & 58.57 & 51.30 & 70.47 & 61.31 \\
& \textbf{Average}   & 79.22  & 75.50 & 74.56  & \textbf{85.68} & 78.73 \\

\midrule

\multirow{3}{*}{Lung \& Heart} 
& AlexNet   & 95.16 & 94.65 & 95.11 & 96.93 & 95.46\\
& ResNet50  & 98.91 & 98.26 & 98.97 & 99.70 & \textbf{98.96} \\
& Vit-B-16  & 91.13 & 90.63  & 90.83 & 93.97 & 91.64\\
& \textbf{Average} & 95.07  & 94.51 & 94.97  & \textbf{96.87} & 95.35 \\

\bottomrule
\end{tabular}
\end{table*}

\subsection{Name That Dataset!}
The exciting area of our paper is the analysis we perform using the models to evaluate the task \textit{Name That Dataset!} on different datasets and other transformations proposed in \cite{zengyin2024bias}. 
This work showed that in natural images, semantic maps and structures dominated the bias, and in fact, local texture and colour played important roles in dataset bias.
They proved this by showing pixel shuffling drastically decreases data classification accuracy.
This is quite different in our analysis due to the nature of medical images. 
X-ray images only contain one channel, so we wish to analyse whether semantic maps and structures also play a role in medical image dataset bias.
As a result, we perform the task on four different dataset transformations: Cropped Images, Semantic Images, Contour Images, and combined Lung and Heart images; examples from each dataset can all be seen in Figure \ref{fig:transforms}.
We make heavy use of the TorchXRayVision \cite{Cohen2022xrv} segmentation model to prepare our transformations.
All accuracies for different model architectures can be seen in Table \ref{tab:combined_results}.

\paragraph{Cropped Images} were created to encompass as much relevant pathological tissue used for diagnosis as possible.
Utilising the segmentation model from TorchXRayVision, a network called PSNet, which builds upon a structure-aware network\cite{Lian2021}, we select the left and right lung, and we combine these into one mask.
Using the mask, we identify where non-zero pixels in the X and Y axes are by producing a boolean array.
We then use this boolean array to find the minimum and maximum coordinates along each axis, which defines a bounding box that tightly surrounds the non-zero area.
We can crop the image to the lung alone by slicing the original image based on the bounding box; we did not add any margin due to the presence of noisy backgrounds. 
Lastly, due to the appearance of machine artefacts, which most often occur in the top two corners, we add two black patches to every image as a transform whilst training for this task.
As the black patches are fixed in size and present in every image, we negate the bias from machine artefacts. 
When evaluating this task, which has the most information of all tasks, we achieve an accuracy of 99.28\% with the ResNet50 model.

\paragraph{Semantic Images} provide highly detailed information on the physiological objects present in each X-ray.
As all humans grow individually and differently, it goes without saying that the semantic segmentation maps of bodily organs should provide rich, classifiable data.
However, as we are dealing with a dataset bias and not an individual bias, the dataset classification accuracy could correlate with hidden demographic data not present in the metadata of the dataset. 
For a human expert, this task increases the difficulty as it would be extremely difficult to label an X-ray image based on the semantic masks alone.
The semantic masks were generated with the same model as previously mentioned; this time, there are no combining masks, and we treat each one separately.
As such, we report an accuracy of 86.61\% with the ResNet50 model, which is well above the chance and only marginally less than the cropped images.

\paragraph{Contour Images} makes use of the semantic segmentation masks previously used to focus on structures that make up the image, more simply a curve that joins all continuous points all the mask boundary.
Using OpenCV \cite{opencv_library}, we locate the boundary points of each mask.
With the same library, we are able to take the boundary points for each mask and draw white lines with a pixel thickness of one.
The resulting image is the contours of each of the 14 masks provided by the segmentation model.
This task is even more challenging compared to the semantic images, as the organ contours appear very similar across images and datasets.
As humans, we would naturally find it very difficult to solve such a task with our eyes alone and no other data available.
Surprisingly, we report a high percentage result of 88.63\% with the ResNet50 model; it appears the models do not struggle with the task as humans would.

\begin{figure*}[t]
\centering
\includegraphics[width=\textwidth]{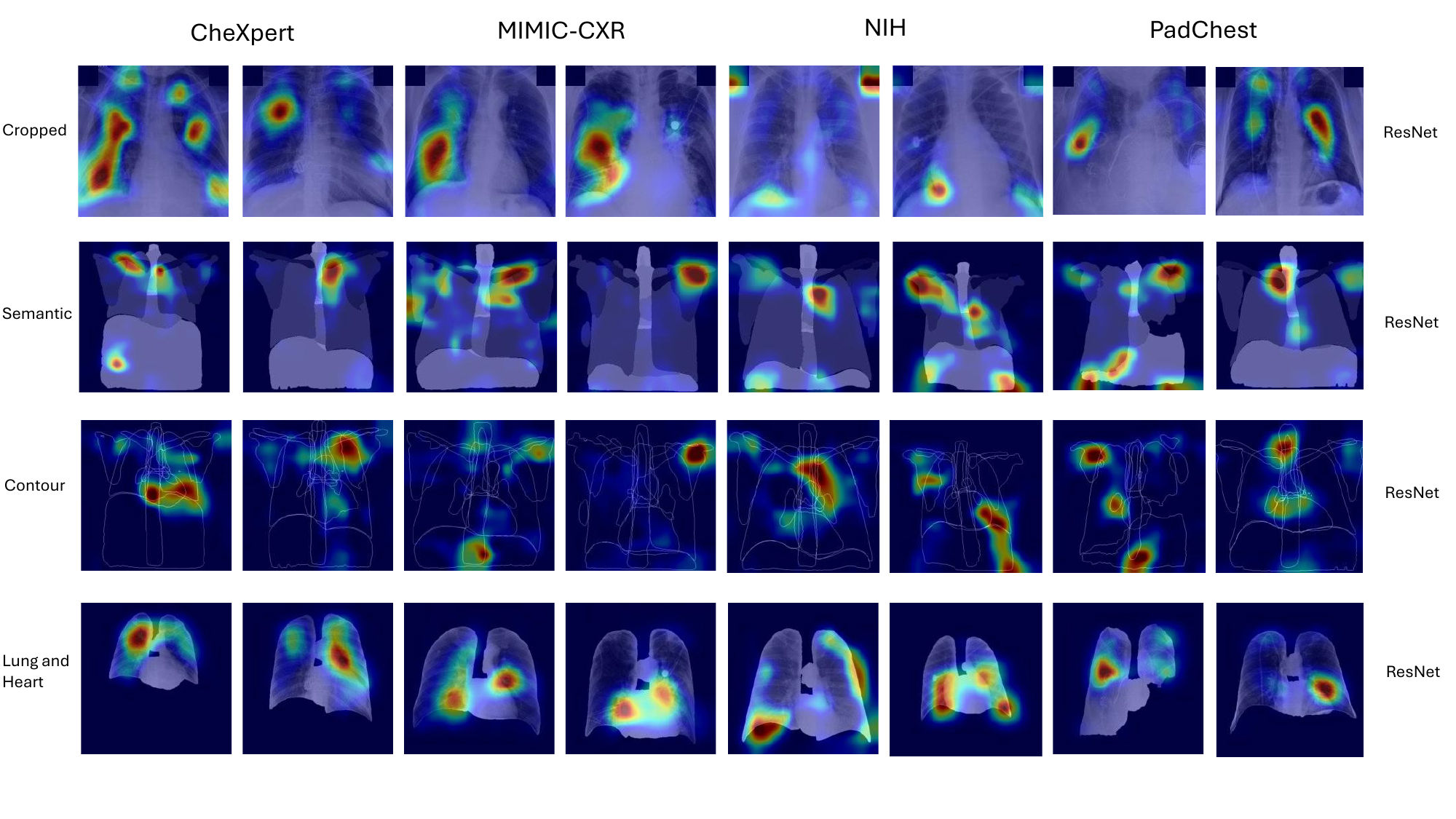}
\caption{Using the same images from Figure \ref{fig:transforms}, we pass each image through its corresponding best-performing model and generate heatmaps based on the last layer to visualise the model’s prediction process. See the supplementary material for more images. In the Cropped and LH images, the model focuses on the lung and heart regions, whereas for the semantic and contour images, the prediction appears more random.}
\label{fig:heatmaps}
\end{figure*}

\paragraph{Lung and Heart Images,} we will refer to as LH images, are generated using the following masks:  left lung, right lung, left Hilus pulmonis, right Hilus pulmonis, heart and the aorta.
With the masks, we concatenate these into a single binary mask and multiply this by the original image, resulting in an image with only the heart and lung tissue.
Since the heart and lungs play a major role in diagnosing diseases within the thoracic region \cite{Speets2006}, we aim to obfuscate all other areas, such as the clavicle. 
Our primary objective is to assess whether models can accurately predict dataset origin based solely on the heart and lungs, and, if so, to investigate why this is possible. 
This task should be less challenging than the previous two, but still more difficult than using cropped images.
The ResNet50 model achieved an accuracy of 98.96\%, only slightly lower than that obtained with the Cropped images.

\paragraph{Experiment settings} for training purposes, we resized all images to the fixed shape of (224,224).
We did not apply data augmentations for two reasons:  the transformations already act as a form of augmentation, and therefore, we wish to see results before further additions.
For the cropped images, we apply the transform of adding black patches, but this is irrelevant in all other tasks.
We perform a Z-score normalisation per image.
We take extreme care to ensure patient-level data exists in either the train or test. 
For \textbf{NIH}, the split is provided; for the other datasets, we use the metadata CSVs to extract the unique list of patients and perform a random 80:20 split, generating two patient lists.
Hyperparameters were chosen to suit each model's best performance, with the AlexNet and ResNet we set the learning rate and weight decay both to 1e-4, whereas the ViT we set to 1e-5.
The AlexNet and ResNet50 both made use of the Cosine scheduler, whereas the ViT we set up a Linear warmup Cosine scheduler, as this has proven best for training ViTs \cite{steiner2022trainvitdataaugmentation}. 
All training uses the cross-entropy loss, the optimiser is fixed to AdamW, and the batch size is also fixed to 64.
Lastly, for merging the datasets, we perform a random permutation of the length of the dataset to generate random indices for each dataset.
We then take a subset from each dataset and concatenate these into one dataset. This is repeated separately for training and validation.
We evaluate the predictions with the F1 score.
All training was performed in PyTorch on a single NVIDIA 4090.


\section{Explaining the Bias}
The results from our experiments have shown impressive and quite alarming results.
Given the high accuracy among our difficult tasks, we need to analyse how our models can achieve such high results.
We break this down into these areas of interest: visualisation, segmentation level analysis, Negating Bias and extra experiments in the form of ablation studies. 
For additional experiments, we use only the ResNet, given that it had the highest performance among all tasks.
\subsection{Bias Visualisation}
As with many of the studies in the related work \ref{sec:related_work}, Grad-CAM \cite{zhou2016learning, selvaraju2017grad} is extremely effective at highlighting key regions in the input image to explain how a model makes its prediction.
We evaluate the best model for each task and produce heatmaps implemented in TorchCam \cite{torcham2020}, shown in Figure \ref{fig:heatmaps}.
We find that the tasks' heatmaps can be split into two findings.
In the Semantic and Contour Heatmaps, the findings appear to be more random when looking at a range of different anatomical structures to make a prediction.
Whereas in the Cropped and LH images, we see much greater emphasis on the lungs. 
Although in the NIH cropped images the model seldom attempts to exploit the black patches, it still primarily focuses on the lung, and this focus becomes even more pronounced when only the lung and heart are present.
This suggests that pixel intensity and texture play an important role in making dataset classification predictions.

\begin{figure}[t]
\centering
\includegraphics[width=\linewidth]{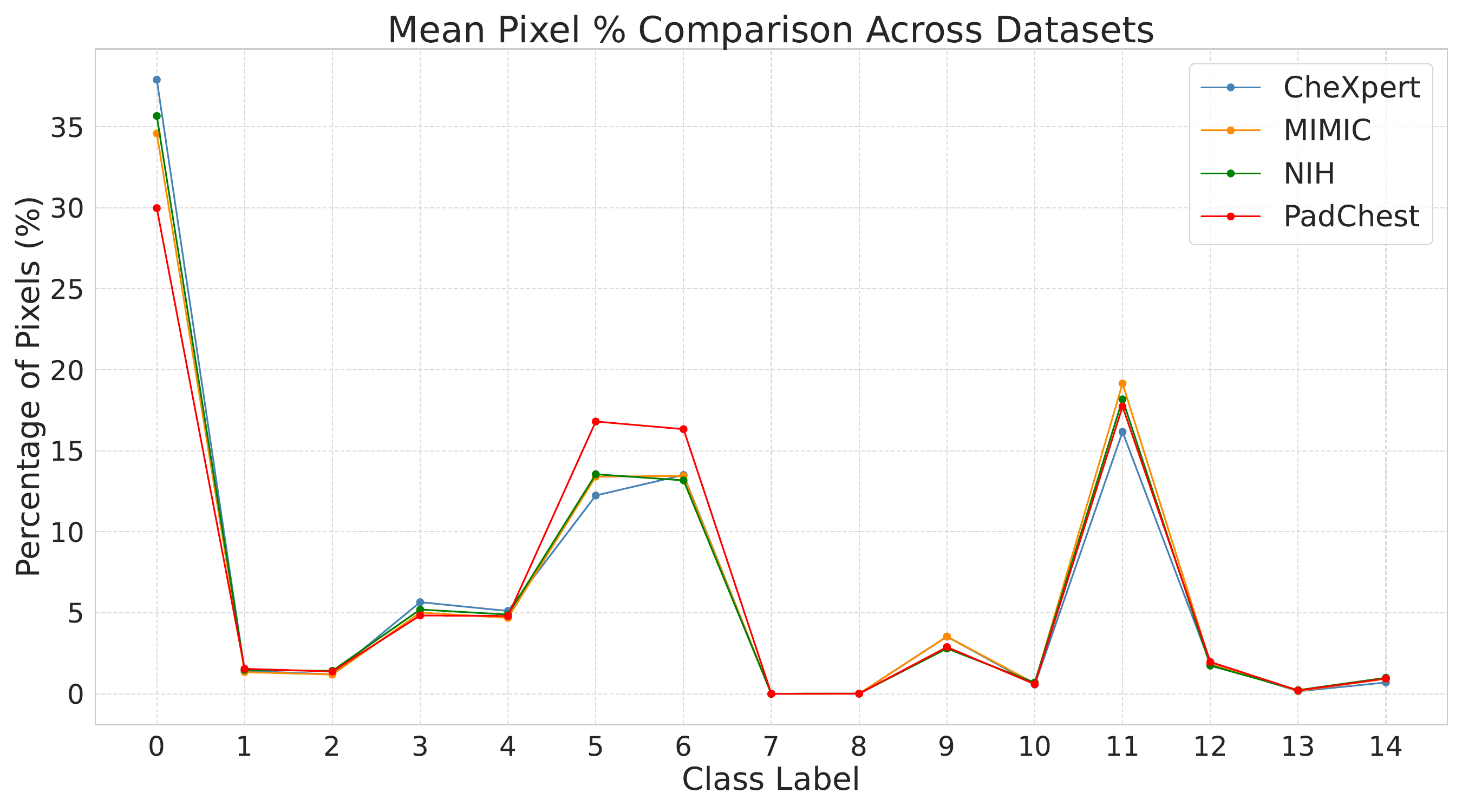}
\caption{We computed the mean pixel percentage per class in each dataset. It is visually hard to distinguish, although classes 0, 5, 6 and 11 do have variations.}
\label{fig:comparison}
\end{figure}
\subsection{Segmentation Level Analysis}

For semantic-level analysis, we compute the proportion of pixels each mask occupies in an image by dividing the mask pixel count by the total pixels. 
The mean across all images per dataset is then reported, as shown in Figure \ref{fig:comparison}.
While many classes have similar average values, certain classes show notable differences—specifically: 0 (Background), 5 (Left Lung), 6 (Right Lung), and 11 (Facies Diaphragmatica).
Based on this, we look closer at the descriptive statistics \ref{tab:descriptive} for these classes to explain the unusually high dataset classification accuracies within the Semantic and Contour image transformation task.
Within the CheXpert dataset, we see that the background count is distinctively much higher than the other datasets, and the PadChest background count is much lower.
We observe that the MIMIC Facies Diaphragmatica is noticeably higher than all other datasets, and the CheXpert is the lowest.
The PadChest dataset has higher counts for both lungs; the CheXpert is the only dataset in which the left lung has a count lower than the right lung.
These counts can explain how models can classify so well within the semantic and contour tasks, but it does not explain how the models perform in the other two tasks.

\begin{table}[t]
\centering
\caption{Dataset Semantic Class Statistics. Although subtle, the differences between the chosen classes are sufficient for a modern neural network to leverage in classifying the dataset origin.}
\label{tab:descriptive}
\setlength{\tabcolsep}{6pt}  
\begin{tabular}{l|c|c|c|c}
\toprule
\textbf{Dataset} & \textbf{Class} & \textbf{Mean} & \textbf{Median} & \textbf{IQR}\\
\midrule
\multirow{3}{4em}{\textbf{CheXpert}} & 0 & 39.90 ± 8.41 & 37.81 & 11.37  \\
 & 5 & 12.23 ± 3.57 & 11.96 & 4.75  \\
& 6 & 13.50 ± 3.68 & 13.26 & 4.38 \\
& 11 & 16.17 ± 6.41 & 16.04 & 8.59 \\
\midrule
\multirow{3}{4em}{\textbf{MIMIC}} & 0 & 34.58 ± 10.26 & 33.51 & 12.41  \\
 & 5 & 13.40 ± 3.77 & 13.26 & 5.09  \\
& 6 & 13.44 ± 3.62 & 13.30 & 4.71 \\
& 11 & 19.15 ± 7.22 & 19.04 & 9.69 \\
\midrule
\multirow{3}{4em}{\textbf{NIH}} & 0 & 35.66 ± 9.44 & 34.47 & 11.68  \\
 & 5 & 13.55 ± 3.88 & 13.53 & 5.25 \\
& 6 & 13.17 ± 3.69 & 13.07 & 4.87 \\
& 11 & 18.18 ± 6.64 & 18.17 & 8.94 \\
\midrule
\multirow{3}{4em}{\textbf{PadChest}} & 0 & 29.98 ± 11.71 & 28.40 & 10.65  \\
 & 5 & 16.80 ± 4.41 & 17.00 & 5.17 \\
& 6 & 16.33 ± 4.31 & 16.38 & 5.09 \\
& 11 & 17.73 ± 5.89 & 17.67 & 7.53 \\
\bottomrule
\end{tabular}
\end{table}

\subsection{Negating Bias}

Our current findings from the visualisation and the segmentation level analysis lead us to believe that the texture and pixel intensity contribute significantly to bias in the cropped and lung and heart only images.
Liu and He \cite{liu2024decade} found that random crops, random augmentation, MixUp and CutMix \cite{szegedy2015going, cubuk2020randaugment, zhang2018mixup, yun2019cutmix} improve dataset classification accuracy.
Since our classes represent the dataset origin, our goal is to reduce classification accuracy—highly predictive image-based features may reflect dataset-specific biases that should be mitigated.
Whilst data augmentation makes it more difficult for models to memorise training images and encourages more generalisable features to be learned, it could also be interpreted that this type of augmentation has actually increased bias due to the increase in dataset classification accuracy.
Heavily influenced by Balestriero et al \cite{balestriero2022effectsregularizationdataaugmentation}, we need to carefully decide on augmentations to help decrease bias in our training.
Balestriero et al prove by empirical analysis that certain augmentations will reduce results in some classes but increase in other classes. 
Given the internal and external findings, we want to investigate the data augmentation techniques: intensity shifting, smoothing, blurring and Mixup.
For augmentations, we implement 13 transforms (see supplementary material) from the MONAI \cite{cardoso2022monaiopensourceframeworkdeep} library that focus on shifting intensity and texture on the cropped and LH images to see if we can decrease bias.
We report in Table \ref{tab:augmentation_ablation} marginal drops in results, and the probability of a transform occurring did not affect results.

\begin{table}[t]
\centering
\caption{Augmentation and Ablation F1 Results. Additional experiments focus on the two image types with pathological tissue.}
\label{tab:augmentation_ablation}
\setlength{\tabcolsep}{10pt}
\renewcommand{\arraystretch}{1.2}
\begin{tabular}{lcc}
\toprule
\textbf{Transform} & \textbf{Cropped Result} & \textbf{LH Result} \\
\midrule
\multicolumn{3}{c}{\textbf{Augmentations}} \\
Transforms (P=0.2) & 98.30 & 97.27 \\
Transforms (P=0.5) & 98.29 & 97.28 \\
MixUp              & 97.62 & 98.21 \\
Combined           & \textbf{95.96} & \textbf{95.06} \\
\midrule
\multicolumn{3}{c}{\textbf{Ablations}} \\
Shuffle Patches    & 97.81 & 96.13 \\
Shuffle Pixels     & 62.52 & 57.81 \\
\bottomrule
\end{tabular}
\end{table}

\subsection{Ablation Studies}
We run two forms of ablation, considering only the cropped images.
In the first ablation, we ran two additional experiments, changing the size of the black patches.
We test setting the black patch size to 50 and 70. 
The results were 99.25 and 99.15, which is only a 0.13 decrease from the original experiment. 
This indicates that the size of the black patch does not influence the model's decisions.
In the second ablation, we also run two more experiments based on the overall image size.
We resize the image down to 32 and 64, scaling the black patch accordingly to 4 and 9.
We present the results 93.67 and 97.35; these results are extremely high given the small resolution. 
These results show that the patterns the model are learning is being scaled down in the bilinear interpolation when resizing the images.
We also conducted two additional ablation studies, evaluating the cropped images in parallel with the LH images.
We test two experiments on both types of image: dividing each image into patches and shuffling them, and shuffling all pixels in the image.
These results in Table \ref{tab:augmentation_ablation} prove our hypothesis that pixel intensity and texture play a vital role in the underlying dataset bias.


\section{Discussion and Conclusion}
This work revisited \textit{Name That Dataset!} within the X-ray modality.
We selected popular large open-source datasets and uniformly sampled frontal X-ray images to conduct this task.
We found that modern networks were able to achieve high dataset classification accuracy results.
Notably, in the semantic and contour tasks, the models achieved results that indicate a capacity for representation and understanding beyond what would be feasible for human experts, given the complexity of these tasks.
Unexpectedly, the contour task achieves higher performance than the semantic task.
The ViT performed worse than CNNs, but this was significantly worse when training on the semantic and contour images.
The reduced performance on semantic and contour images may stem not only from the inherent training instability of ViTs, but also from the fact that these image types may not provide the structural cues that self-attention mechanisms rely on for effective pattern recognition.
Unlike \cite{zengyin2024bias}, where semantic maps and structures were the main sources of bias, our findings indicate that these factors are largely irrelevant. 
Instead, dataset bias is primarily driven by pixel intensity and texture rather than anatomical structure.
The slightly worse performance in cropped and LH images suggests that ViTs show less dataset bias towards pixel texture, which is further backed in studies \cite{geirhos2022imagenettrainedcnnsbiasedtexture, naseer2021intriguingpropertiesvisiontransformers, park2022visiontransformerswork}.
Semantic analysis shows the class percentages are similar, and when looking at visualisations, the cropped and LH models focus heavily on the lung and heart tissue.
This is supported by additional experiments using a large number of augmentations, where a small decrease in accuracy was reported. 
Regardless of how much we altered the images, the models were still able to achieve high accuracy.
Alongside this, ablations showed that shapes in the images did not have a large effect on dataset classification accuracy, especially when patch shuffling or pixel shuffling.
More research is needed to analyse pixel intensity and texture across overlapping disease classes, producing a class-by-class analysis for each dataset. 
Disease classification models may be affected if significant differences are present.
This could cause concern for foundation models trained on large datasets; if a disease class exhibits only a narrow range of pixel intensities and textures, the model may fail to correctly classify outliers, especially in tasks like long-tailed classification \cite{holste2022long}.
Future work will examine more datasets and modalities, focusing on mitigating intensity and texture-based biases.

\section{Acknowledgements}
We would like to give thanks to Dr. Lorenzo Brigato and Shelley Zixin Shu for the useful discussions in this work.
Calculations were performed on UBELIX (https://www.id.unibe.ch/hpc), the HPC cluster at the University of Bern.

{
    \small
    \bibliographystyle{ieeenat_fullname}
    \bibliography{main}
}

\clearpage
\setcounter{page}{1}

\maketitlesupplementary
\begin{strip}
\centering
\includegraphics[width=\textwidth]{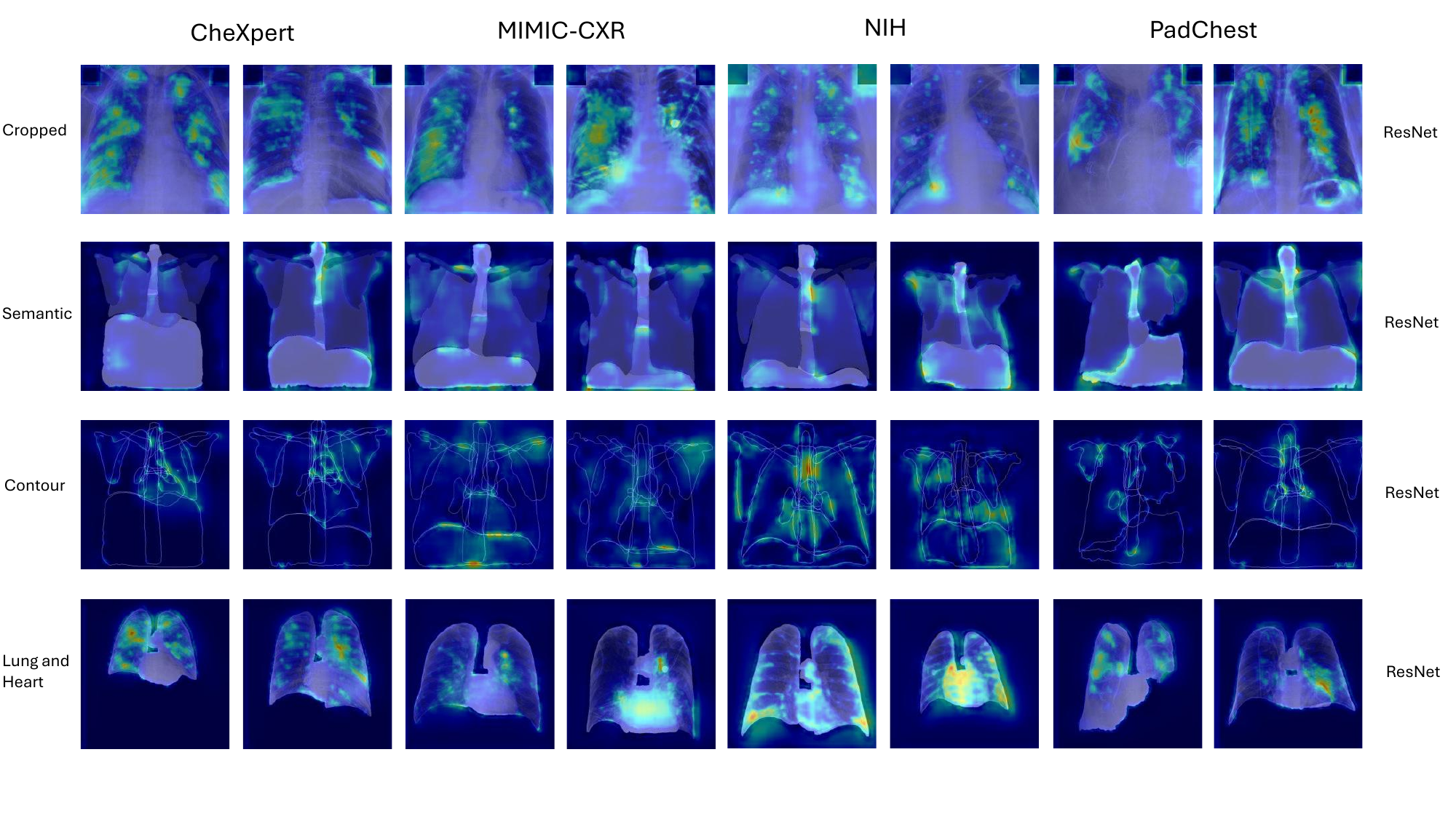}
\captionof{figure}{Likewise, as in the first heatmap figure, we pass the same images through their corresponding best-performing model and generate heatmaps based on all layers to visualise the model’s prediction process.}
\label{fig:heatmaps_averages}
\end{strip}

\section{Average Heatmaps}
\label{sec:average_heatmaps}
In the previous heatmap figure, the Grad-CAM was generated using only the final activation layer.
As a comparison, we now generate heatmaps by averaging the Grad-CAMs computed across all model layers of the best-performing model.
Looking at Figure \ref{fig:heatmaps_averages}, while the heatmaps for the cropped and LH images remain relatively similar, the semantic and contour images provide new insights.
These reveal a broader range of regions the model attends to, and the attention patterns appear less random than those based on the final layer alone.
The model seems to be analysing various anatomical structures to determine the dataset of origin.
This suggests a form of semantic-level analysis, similar to what we computed, where the model approximates organ structures and assesses which dataset they most likely correspond to.
\vspace{1em}
\section{Transformations}
\label{sec:Transforms}

We utilise the \textbf{MONAI library} to implement 13 carefully selected augmentation transformations designed to enhance model robustness through controlled variations in pixel intensity and texture. Our experiments evaluate these transformations using two distinct application probabilities: $P = 0.2$ (conservative) and $P = 0.5$ (aggressive).

\subsection*{Augmentation Transformations}
The complete set of transformations includes:

\subsubsection*{Intensity Modifications}
\begin{itemize}[leftmargin=*,nosep]
    \item \textbf{Gaussian Noise} \\
    \texttt{RandGaussianNoise} (prob = $P$)
    
    \item \textbf{Intensity Shifting} \\
    \texttt{RandShiftIntensity} (offset = 0.1, prob = 0.5) \\
    \texttt{RandStdShiftIntensity} (factor = 0.1, prob = 0.5)
    
    \item \textbf{Intensity Scaling} \\
    \texttt{RandScaleIntensity} (factor = 0.1, prob = 0.5) \\
    \texttt{RandScaleIntensityFixedMean} (factor = 0.1, prob = 0.5)
    
    \item \textbf{Contrast Adjustment} \\
    \texttt{RandAdjustContrast} (prob = 0.5)
\end{itemize}

\subsubsection*{Spatial \& Texture Modifications}
\begin{itemize}[leftmargin=*,nosep]
    \item \textbf{Smoothing Filters} \\
    \texttt{SavitzkyGolaySmooth} (window\_length = 5, order = 2, prob = 0.5) \\
    \texttt{RandGaussianSmooth} ($\sigma$ = 1.0, prob = 0.5) \\
    \texttt{MedianSmooth} (radius = 1, prob = 0.5)
    
    \item \textbf{Sharpening} \\
    \texttt{RandGaussianSharpen} (prob = 0.5)
    
    \item \textbf{Non-linear Transforms} \\
    \texttt{RandHistogramShift} (control\_points = 10, prob = 0.5)
\end{itemize}

\subsubsection*{Structural Perturbations}
\begin{itemize}[leftmargin=*,nosep]
    \item \textbf{Dropout \& Shuffling} \\
    \texttt{RandCoarseDropout} (holes = 5, size = (32,32), prob = 0.5) \\
    \texttt{RandCoarseShuffle} (holes = 5, size = (32,32), max\_holes = 10, prob = 0.5)
\end{itemize}

We aim to preserve the existing transform pipeline used during training and simply insert a MONAI transform within it. 
Since MONAI expects input as NumPy arrays with a single channel, we add a custom transform after resizing the PIL image to convert it to a NumPy array and append a channel dimension. After applying the MONAI transforms, we convert the output (a MONAI MetaTensor) back to a NumPy array and apply another custom transform to remove the channel dimension. 
This ensures compatibility while keeping the rest of the transformation pipeline unchanged.

\end{document}